\documentclass[runningheads]{llncs}
\usepackage{graphicx}

\usepackage{comment}
\usepackage{amsmath,amssymb} 
\usepackage{color}
\usepackage{hyperref}
\usepackage{tikz}

\usepackage[accsupp]{axessibility}  

\usepackage{booktabs} 
\usepackage{multirow}
\usepackage{colortbl}
\usepackage{multirow, makecell}
\usepackage{enumitem}
\usepackage{graphicx}
\usepackage{animate}
\usepackage{bm}
\usepackage{stackengine}

\usepackage[capitalize]{cleveref}
\crefname{section}{Sec.}{Secs.}
\Crefname{section}{Section}{Sections}
\Crefname{table}{Table}{Tables}
\crefname{table}{Tab.}{Tabs.}
\Crefname{figure}{Figure}{Figures}
\crefname{figure}{Fig.}{Figs.}

\newcommand{\onedot}{.\null}
\def\eg{\emph{e.g}\onedot} 
\def\ie{\emph{i.e}\onedot}

\begin{document}
\pagestyle{headings}
\mainmatter
\def\ECCVSubNumber{2786}  

\title{CompNVS: Novel View Synthesis\\with Scene Completion}


\titlerunning{CompNVS: Novel View Synthesis with Scene Completion}
\author{
Zuoyue Li\inst{1} \and
Tianxing Fan\inst{2} \and
Zhenqiang Li\inst{3} \and
Zhaopeng Cui\inst{2} \and
\\Yoichi Sato\inst{3} \and
Marc Pollefeys\inst{1,4} \and
Martin R. Oswald\inst{1,5}
}
\authorrunning{Z. Li et al.}
\institute{
$^\text{1\enspace}$ETH Z\"urich \quad
$^\text{2\enspace}$Zhejiang University \quad
$^\text{3\enspace}$The University of Tokyo \quad
\\$^\text{4\enspace}$Microsoft \quad
$^\text{5\enspace}$University of Amsterdam
}

\newcommand{\OURS}{CompNVS}
\newcommand{\USEADOBE}{Please use \textbf{Adobe Reader} / \textbf{KDE Okular} to see \textit{\textbf{animations}}.}
\newcommand{\ANIMATION}{(\textit{animations})}
\newcommand{\STARTFRAME}{00} 
\newcommand{\ENDFRAME}{14} 
\newcommand{\POSTERFRAME}{07} 

\newcommand{\best}[1]{\textbf{#1}}
\newcommand{\boldparagraph}[1]{\noindent{\bf #1} }
\newcommand{\italicparagraph}[1]{\noindent{\it #1} }
\newcommand{\zuoyue}[1]{{\textcolor{red}{#1}}} 
\newcommand{\ftx}[1]{{\textcolor{green}{#1}}} 
\newcommand{\lzq}[1]{\textcolor{orange}{#1}} 
\newcommand{\zp}[1]{{\textcolor{blue}{#1}}} 
\newcommand{\mo}[1]{{\textcolor{orange}{#1}}} 
\newcommand{\todo}[1]{\textcolor{red}{TODO: #1}} 

\newcolumntype{L}[1]{>{\raggedright\let\newline\\\arraybackslash\hspace{0pt}}m{#1}}
\newcolumntype{C}[1]{>{\centering\let\newline\\\arraybackslash\hspace{0pt}}m{#1}}
\newcolumntype{R}[1]{>{\raggedleft\let\newline\\\arraybackslash\hspace{0pt}}m{#1}}

\maketitle
\begin{abstract}
We introduce a scalable framework for novel view synthesis from RGB-D images with largely incomplete scene coverage.
While generative neural approaches have demonstrated spectacular results on 2D images, they have not yet achieved similar photorealistic results in combination with scene completion where a spatial 3D scene understanding is essential.
To this end, we propose a generative pipeline performing on a sparse grid-based neural scene representation to complete unobserved scene parts via a learned distribution of scenes in a 2.5D-3D-2.5D manner.
We process encoded image features in 3D space with a geometry completion network and a subsequent texture inpainting network to extrapolate the missing area.
Photorealistic image sequences can be finally obtained via consistency-relevant differentiable rendering.
Comprehensive experiments show that the graphical outputs of our method outperform the state of the art, especially within unobserved scene parts.

\keywords{novel view synthesis, scene completion, scene representation learning, 3D inpainting/extrapolation, 3D-aware generative modeling}
\end{abstract}
\section{Introduction}
Recent advancements in 3D reconstruction and differentiable rendering have driven impressive progress in novel view synthesis (NVS) from single or multiple images.
With such technology captured scenes can be easily virtually explored.
However, with increasing scene complexity, the capturing effort for a complete scene increases substantially, and in practice captured scenes almost always contain holes due to occlusions and incomplete captures.
Therefore, to synthesize such scenes immersively, the ability to extrapolate visual content into unseen areas in a spatially consistent manner is utterly important.

While GAN-based 2D inpainting approaches~\cite{yu2018generative,yu2018free} can tackle the problem, they typically have trouble learning 3D correspondences and as a result, synthesized videos lack spatial and temporal consistency.
More recent single-image NVS methods~\cite{whiles2020synsin,rockwell2021pixelsynth} combine 3D scene representations with generative modeling and become more 3D-aware.
However, the multi-view inconsistency still occurs in the rendered sequences with large perspective changes due to the completion in 2D space.
Grid-based 3D approaches can achieve better consistency, but may still fail when the input scene is incomplete due to sparse views~\cite{dai2021spsg}.

\begin{figure}[!t]
    \raggedleft
    \scriptsize
    \newcommand{\sz}{0.165\linewidth} 

    \stackinset{r}{0.025\linewidth}{t}{-0.03\textwidth}
        {\includegraphics[width=0.14\linewidth]{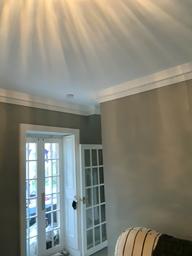}}
        {\includegraphics[width=0.14\linewidth]{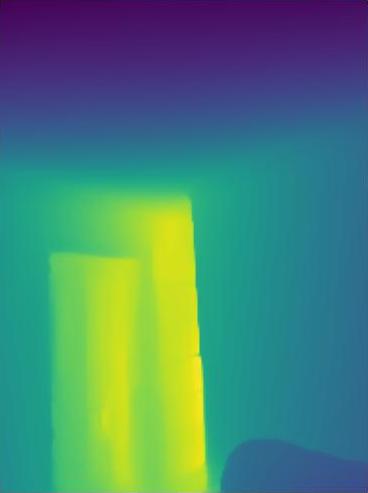}}
    \stackinset{r}{0.025\linewidth}{t}{-0.03\textwidth}
        {\includegraphics[width=0.14\linewidth]{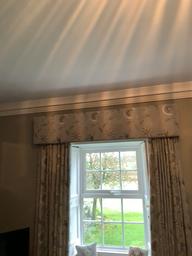}}
        {\includegraphics[width=0.14\linewidth]{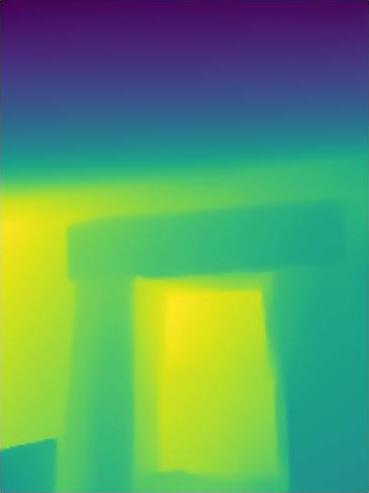}}
    \noindent\animategraphics[autoplay,loop,width=\sz,palindrome,poster=\POSTERFRAME]{5}{images/Arkit_Input/42446156_79162312_79160313_79163312/}{\STARTFRAME}{\ENDFRAME}
    \noindent\animategraphics[autoplay,loop,width=\sz,palindrome,poster=\POSTERFRAME]{5}{images/Arkit_PixelNerf/42446156_79162312_79160313_79163312/}{\STARTFRAME}{\ENDFRAME}
    \noindent\animategraphics[autoplay,loop,width=\sz,palindrome,poster=\POSTERFRAME]{5}{images/Arkit_PixelSynth/42446156_79162312_79160313_79163312/}{\STARTFRAME}{\ENDFRAME}
    \noindent\animategraphics[autoplay,loop,width=\sz,palindrome,poster=\POSTERFRAME]{5}{images/Arkit_Ours/42446156_79162312_79160313_79163312/}{\STARTFRAME}{\ENDFRAME}
    \\
    \setlength{\tabcolsep}{1pt}
    \begin{tabular}{C{0.315\linewidth}C{\sz}C{\sz}C{\sz}C{\sz}}
    Posed partial observations & No completion  &  pixelNeRF \cite{yu2021pixelnerf} & PixelSynth \cite{rockwell2021pixelsynth} & \textbf{Ours}
    \end{tabular}
    \caption{\textbf{Examplary input and output of our method.} Our approach combines generative modeling with a sparse 3D feature representation to complete unobserved scene parts plausibly. \USEADOBE}
    \label{fig:teaser}

\end{figure}

In recent years, we have also seen the introduction of neural implicit representations~\cite{mildenhall2020nerf,liu2020nsvf} which further improved the (differentiable) rendering standards. 
These representations no longer strive for explicit 3D data structures but to ``store'' a single scene implicitly in the network, without being limited to a specific spatial resolution anymore.
As a result, the quality of rendering images for the novel views is significantly improved and they can meanwhile hold very good multi-view consistency.
However, these methods focus more on 3D reconstruction or NVS from multiple images with good coverage.
Before inference for new query views, they have to optimize the corresponding scene individually, using its hundreds of observations.
This kind of test-time optimization generally may not learn the priors across different scenes, and it is also difficult to directly use them for (conditional) scene generation or completion.
There are also several applications of implicit representation combined with differentiable rendering in the field of scene or object generation.
Most existing approaches~\cite{ngu2019hologan,niemeyer2020giraffe,chan2020pigan,devries2021gsn,ngu2020blockgan} focus more on sampling from latent noise, and rarely involve conditions on some input.
And they still rely heavily on the 2D network to ensure the generation quality, which may break the multi-view consistency from the 3D part. 

In this project, we tackle the problem of 3D scene generation from limited partial observations. 
We present an approach that takes as input a partially scanned point cloud from a single or multiple posed RGB-D images and outputs novel synthesized views of a completed 3D model in a spatially consistent manner. 
Specifically, we perform geometry and texture synthesis directly in 3D space utilizing neural implicit representations with an explicit voxel grid.
Exemplary results of our work are shown in \cref{fig:teaser}.
We hope that such an architecture provides stronger generation abilities, easier learns 3D priors from multiple scenes, and can thus generate high-quality multi-view consistent images via differentiable rendering, and benefit scene editing, or NVS from sparse observations.

In sum, we make the following contributions:
(1) We propose a scalable NVS framework capable of generative modeling to complete larger unobserved scene parts with plausible and spatially consistent content.
(2) We adapt NSVF~\cite{liu2020nsvf} to a feed-forward version, in which the scene embeddings can be directly obtained by an encoder without test-time optimization.
(3) We demonstrate superior experiment results on both synthetic and real-world datasets.

\section{Related Work}

Our work is at the cross-section of NVS, scene completion, and 3D generative modeling, which all have been studied with various scene representations implying crucial method properties.
In the following, we discuss related works according to their underlying representations.

\boldparagraph{Explicit scene representations.}
Early 3D generative methods such as HoloGAN~\cite{ngu2019hologan} or BlockGAN~\cite{ngu2020blockgan} use the 3D representation of the voxel grid. The generation is performed in a coarse feature volume, which is further projected to each frame using the query camera poses, followed by upsampling and super-resolution through the 2D generation network to get the final result. 
The key idea is that the projected features naturally hold temporal consistency, but it is difficult to get high-quality results because of the limitation of the 3D volume resolution. 
Such a constraint also applies to scene completion methods like SPSG~\cite{dai2021spsg}, which performs generation directly on a TSDF volume. 
On the other hand, in the field of single-image NVS, SynSin~\cite{whiles2020synsin} and PixelSynth~\cite{rockwell2021pixelsynth} use point cloud representation, which is also adopted by the recent works of cross-view generation Sat2Vid~\cite{li2021sat2vid} and video translation WC-Vid2Vid~\cite{mallya2020wcvid2vid} to encourage temporal consistency of the output video. 
The projection of point cloud is often finer than volumes, but the rendered frame usually has ``holes'', which also require a subsequent 2D network to make the completion, where the temporal consistency is hard to control especially when the ``holes'' are large.
We see that either voxel grid or point cloud can obtain an initial multi-view consistency from a 3D explicit representation, however, such consistency could be easily destroyed by the 2D CNNs, which may still result in unsatisfactory video quality.

\boldparagraph{Implicit scene representations.}
The emergence of NeRF~\cite{mildenhall2020nerf} allows scene information to be no longer stored explicitly like voxel grids or point clouds, but to map 3D coordinates (and view directions) to RGB/alpha values through a multi-layer perceptron (MLP), making it not limited by resolution. 
By introducing implicit representation to the differentiable rendering techniques, it can render high-quality photo-realistic images of an object or a scene from unseen points of view and can generate the video with very well temporal consistency. 
However, NeRF optimization requires hundreds of images of the same scene or object. 
For different scenarios, different radiance fields are required to be trained individually. 
Such test-time optimization settings and the implicit representation make it inherently difficult for the generation. 
Thus, how to combine GAN techniques and implicit representations would be an attractive direction. 

\noindent
GRAF~\cite{schwarz2020graf} is a pioneer work in the field of \textbf{generative NeRF}. It simply concatenates two global latent vectors representing the object shape and appearance respectively with the original input positions of MLP, which may not be powerful enough to handle complicated scenes. 
As NeRF’s rendering speed is slow, it samples patches for the discriminator to speed-up training which may not work for a high-resolution generation. 
piGAN~\cite{chan2020pigan} further improves the visual quality by introducing StyleGAN-like~\cite{karras2019stylegan} mapping and synthesis network with SIREN~\cite{sitzmann2020siren} activation, but is still far away from 2D GANs and the training and inference are slow. 
GIRAFFE~\cite{niemeyer2020giraffe} proposes a compositional NeRF followed by a 2D upsampler. The basic idea is similar to GRAF, but it further decomposes the 3D scene representation into several controllable local feature fields. 
It sacrifices the resolution of rendering in exchange for speed and then uses 2D CNN for super-resolution, which further impedes the multi-view consistency. 
In sum, direct generative NeRF training like GRAF and piGAN may suffer from low computation speed and is only acceptable at low resolution. 
Works like GIRAFFE using 2D CNNs can improve the visual quality but again have the problem of the multi-view consistency. 
More importantly, these works sample scenes from learned distributions, without conditioning on the given observations.

\begin{figure*}[!t]
	\centering
	\includegraphics[width=0.91\linewidth]{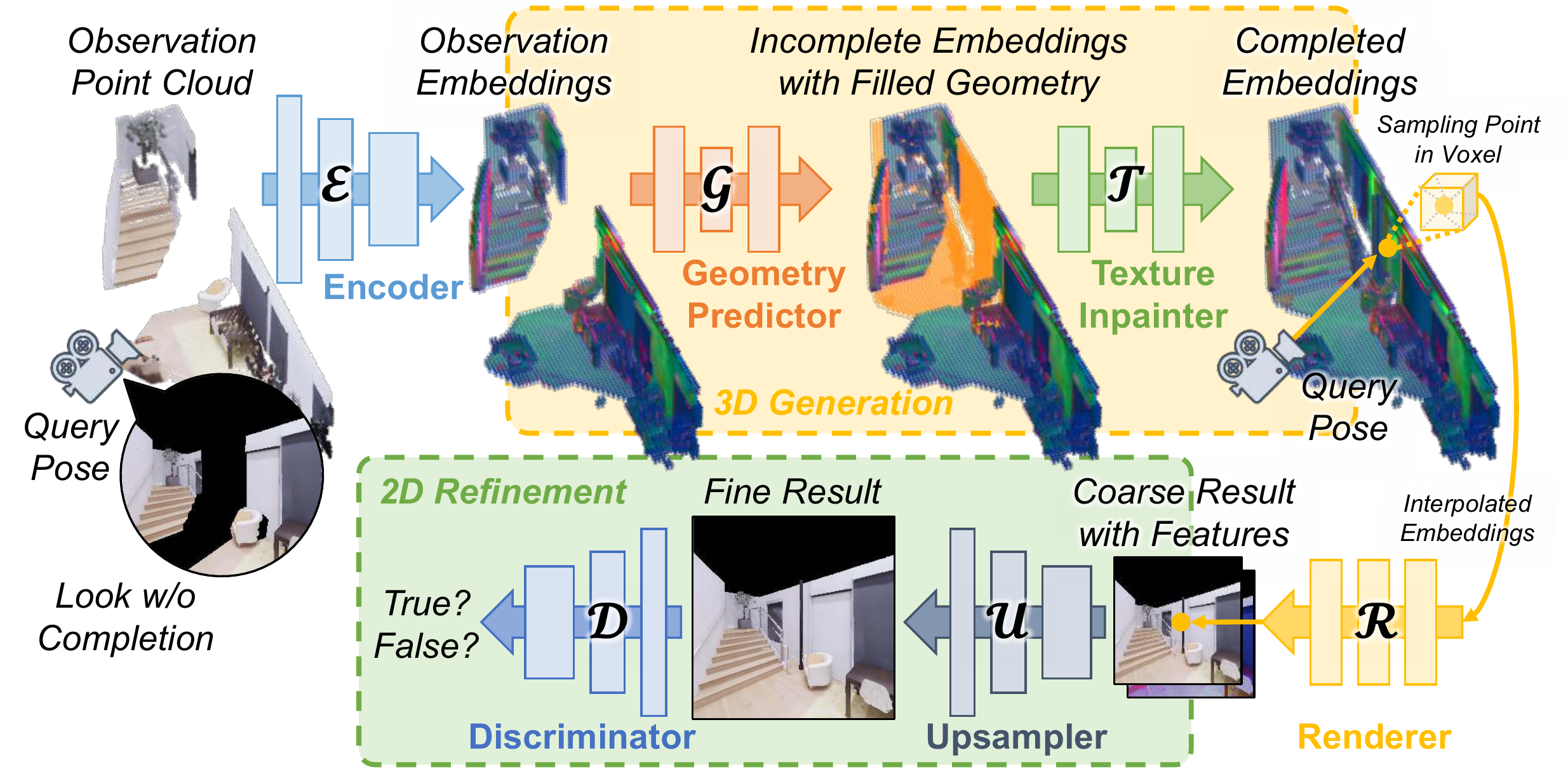}
	\caption{
	\textbf{Network architecture overview.} Our pipeline generates a spatially consistent 3D representation for scene completion and novel view synthesis. The pipeline can be divided into four major steps: input feature encoding, generative scene completion (geometry \& texture), differentiable rendering, and 2D upsampling and refinement.
	}
	\label{fig:pipeline}
\end{figure*}

\boldparagraph{Hybrid scene representations.}
Based on the implicit radiance field, NSVF~\cite{liu2020nsvf} further introduces an explicit representation of a sparse voxel grid to store the information of local neighborhoods, thus solving the problem of insufficient expression ability of MLP in NeRF. 
With the explicitly saved local scene embeddings, this hybrid approach can make the scene more scalable and no longer limited to the object level, and the generated results are more refined than NeRF. 
However, NSVF does not have a generative setting and still requires hundreds of images for test-time optimization. 
GSN~\cite{devries2021gsn} has similar ideas to NSVF and incorporates the GAN setting. 
It generates a 2D feature layout of a scene (vertical top view) from a latent code first, on which the rendering rays can thus extract features, followed by a general NeRF pipeline. 
Another work, pixelNeRF~\cite{yu2021pixelnerf} can learn to predict a NeRF from a few posed images, allowing it to generate plausible novel views without test-time optimization. 
It assigns features from the input views to the sampling points via projection when rendering the target view, making the generation more applicable and image conditioned. 
However, such feature acquisition is naive and the points' spatial relationships are not well established, making the geometry difficult to learn, and further leading to blurry results.

\section{Method}

%
\boldparagraph{Pipeline overview.}
A central goal of our work within the NVS setting is the completion of both the geometry and the texture of larger missing scene parts when observed from a novel query viewpoint. 
In contrast to 2D inpainting approaches, we aim to generate a completed 3D scene representation with texture information, which can be used to render videos with ensured spatial and temporal consistency. 
To this end, we propose the pipeline illustrated in \cref{fig:pipeline} which takes as input an incomplete point cloud obtained by combining all known RGB-D input views. 
The input point cloud is encoded into a sparse voxel grid of 3D features, followed by two predictor networks that complete missing geometry by means of voxel occupancies and texture inpainting subsequently. 
Finally, for a given camera pose or trajectory, a single 2D image or video can be obtained via implicit differentiable rendering followed by a 2D upsampling and refinement module. 
We detail each component of the pipeline as follows. 

%
\boldparagraph{Input and output.}
The input to our method is a point cloud of a partially scanned scene and is reconstructed from given posed RGB-D images. 
The output is the synthesized image or video for a given query view or camera trajectory.

\boldparagraph{Point cloud encoder.}
Given a colored point cloud $(\bm{P}, \bm{C})$ with $\bm{P}\in\mathbb{R}^{N\times3}$ denoting the 3D positions of $N$ points and $\bm{C}\in\mathbb{R}^{N\times3}$ representing their colors, we first employ a ResNet~\cite{he2016resnet} module $\bm{\mathcal{E}}$ based on a sparse voxel representation to encode them into $M$ vertex embeddings $(\bm{V}, \bm{F})=\bm{\mathcal{E}}(\bm{P}, \bm{C})$.
Here the $M$ embeddings $\bm{F}\in\mathbb{R}^{M\times{d}}$ with $d$ dimensional features correspond to all vertices of the occupied voxels, and $\bm{V}\in\mathbb{R}^{M\times3}$ records their coordinates.

\boldparagraph{Geometry predictor.}
In a second step we fill the geometry of the incomplete input scene. 
We adopt a geometry predictor $\bm{\mathcal{G}}$, which yields the filled scene geometry $\widetilde{\bm{V}}\in\mathbb{R}^{(M+m)\times3}$ with $m$ new occupied voxel vertices conditioned on the given $\bm{V}$, \ie, $\widetilde{\bm{V}} = \bm{\mathcal{G}}(\bm{V})$.
Please note that geometry filling is not directly based on the vertex coordinates but based on its associated occupied voxel grid.
Here we omit the step of transformation between a voxel and its indices for simplicity.
Also, $\bm{F}$ is not used in the geometry completion step.
We implement the geometry predictor $\bm{\mathcal{G}}$ as a sparse voxel U-Net~\cite{ronneberger2015unet} model with generative sparse convolutions~\cite{gwak2020gsdn} that can generate new coordinates.

\boldparagraph{Texture inpainter.}
The newly added vertices from the previous step do not hold color embeddings yet and are required to be filled with the necessary texture information.
We predict embeddings for the $m$ vertices by composing a texture inpainter $\bm{\mathcal{T}}$ which predicts the filled embeddings $\widetilde{\bm{F}}\in\mathbb{R}^{(M+m)\times d}$ by taking as input the intial voxel embeddings $\bm{F}_0=\bm{F}\oplus{\bm{0}^{m\times{d}}} \in\mathbb{R}^{(M+m)\times{d}}$ which are formed by padding zeros on $\bm{F}$ for the $m$ new vertices, here $\oplus$ denotes the concatenation operation.
From the visibility information of the input we also indicate whether a vertex needs inpainting or not, so we have $\bm{v}={\bm{0}^M}\oplus{\bm{1}^m}\in\mathbb{R}^{(M+m)} $ and $\widetilde{\bm{F}}=\bm{\mathcal{T}}(\widetilde{\bm{V}},\bm{F}_0,\bm{v})$.
We again employ a U-Net architecture with general sparse convolutions which only operate across the occupied locations.

\boldparagraph{Differentiable renderer.}
In this step, we leverage the rendering pipeline of NSVF to generate a 2D image for the query view based on the explicit sparse voxel representation together with the implicit MLPs.
We follow NSVF to get the feature of a sampling point inside one voxel (illustrated as a yellow cube in \cref{fig:pipeline}) by trilinearly interpolating the embeddings at its 8 vertices.
In the original NSVF work, the features representing RGB and geometry are entangled in the same embedding, which is decoded to RGB and alpha value through the same MLP with two different tail branches.
In contrast, we directly split the embeddings into two groups (1:3) in the feature dimension, and use two independent MLPs to decode the alpha value and RGB respectively, so as to make the embeddings disentangled.
Another difference is that for each pixel, along its corresponding ray, we not only aggregate RGB values, but also compute a feature by assembling the embeddings based on the calculated alpha value. 
This feature will be used for subsequent refinement.
In short, the renderer $\bm{\mathcal{R}}$ takes as input the inpainted voxel embeddings $(\widetilde{\bm{V}},\widetilde{\bm{F}})$, the query pose $\bm{Q}$ and outputs an image with features in half resolution, \ie, $\tilde{\bm{I}} = \mathcal{R}(\tilde{\bm{V}}, \tilde{\bm{F}}, \bm{Q})\in\mathbb{R}^{\frac{H}{2}\times\frac{W}{2}\times(3+d)}$.

\boldparagraph{Refinement module.}
Inspired by Sat2Vid~\cite{li2021sat2vid} we introduce a quality-enhancing refinement module after the rendering, which consists of a lightweight upsampler and a discriminator used during training.
Generally, the paired discriminator improves the ability of generating visually plausible images through adversarial learning.
The training with the discriminator $\bm{\mathcal{D}}$ requires the images to be fully rendered. 
However, directly rendering the full-resolution image with NSVF is highly time-consuming. 
Additionally, the generated images tend to be blurry for larger voxel sizes.
To overcome these challenges, our pipeline first renders a coarse half-resolution image and then uses an upsampler to double the resolution, \ie, $\bm{I}=\bm{\mathcal{U}}(\tilde{\bm{I}})\in\mathbb{R}^{H\times{W}\times{3}}$, which greatly shortens the rendering time.
Please note that the upsampler is designed lightweight in order to not excessively destroy the temporal consistency obtained in rendering phase.

\begin{figure*}[t]
	\centering
	\includegraphics[width=\linewidth]{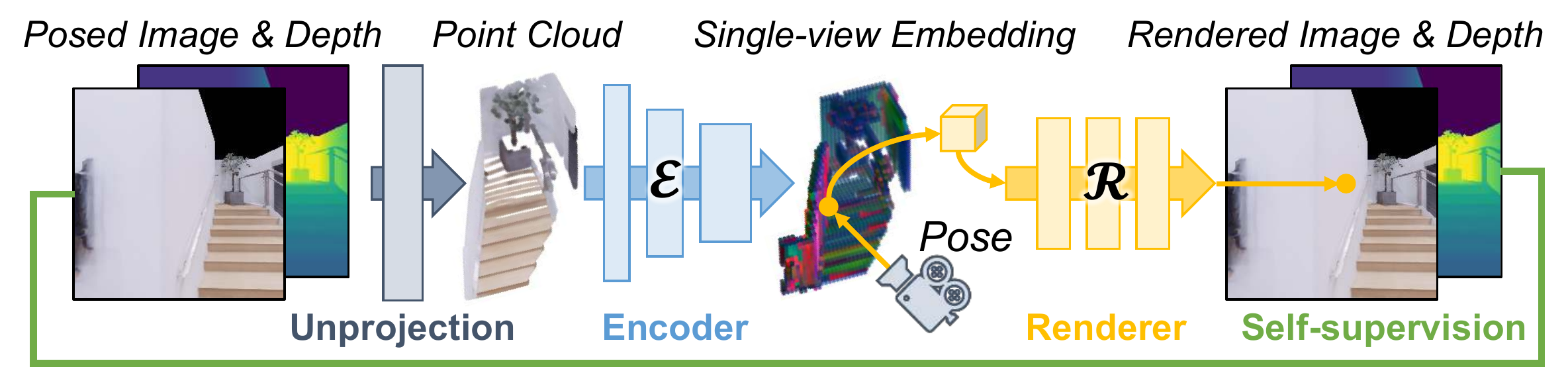}
	\caption{
	\textbf{The first step of training.} We first train the encoder $\bm{\mathcal{E}}$ and the renderer $\bm{\mathcal{R}}$ jointly in a self-supervision way by forming them as a 2.5D-3D-2.5D autoencoder.
	}
	\label{fig:ae}
\end{figure*}

\boldparagraph{Network training.}
The modules are trained step-by-step in four phases. 

\italicparagraph{Step 1.}
We first train the encoder $\bm{\mathcal{E}}$ jointly with the implicit renderer $\bm{\mathcal{R}}$ (MLPs), which forms a structure similar to an autoencoder but in a 2.5D-3D-2.5D manner. 
This procedure is illustrated in \cref{fig:ae}.
Different from NSVF which uses images to optimize the sparse voxel embeddings and the implicit renderer, we use an encoder to directly generate embeddings, and the parameters of the implicit renderer are also shared across scenes.
Thus, a test-time optimization is not required here.
The self-supervision loss is calculated on sampled rays during training instead of the entire image, using both RGB and depth values for training.
To achieve good rendering quality, $\bm{\mathcal{E}}$ here is required to have the ability to encode both geometry and texture information at the same time.

\italicparagraph{Step 2.}
Next we train the geometry predictor $\bm{\mathcal{G}}$ separately.
The decoder of $\bm{\mathcal{G}}$ consists of several coarse-to-fine generative transposed convolution layers~\cite{gwak2020gsdn}, where the binary cross-entropy loss is used for the occupancy prediction.

\italicparagraph{Step 3.}
Then we pre-train the texture inpainter $\bm{\mathcal{T}}$.
We use $\bm{\mathcal{E}}$ trained on the first step to generate necessary paired incomplete and complete embeddings for training $\bm{\mathcal{T}}$.
During pre-training, the ground-truth geometry for the missing part is used, and the feature loss is only calculated in this area, \ie, the orange part in the ``incomplete embeddings with filled geometry'' in \cref{fig:pipeline}.

\italicparagraph{Step 4.}
The final step is to train the 2D refinement module $\bm{\mathcal{U}}$ and $\bm{\mathcal{D}}$ and possibly refine $\bm{\mathcal{F}}$ to adapt it to the predicted geometry in the full pipeline.
We fix the parameters of the other modules during training.
Here the losses are only calculated on the 2D part, \ie, the RGB and depth losses for the half-resolution coarse results, as well as the general GAN losses for the upsampler and discriminator.
There is no embedding supervision for $\bm{\mathcal{F}}$ in this step.

\section{Experiments}

\subsection{Configuration}

\boldparagraph{Datasets.}
Experiments are conducted on both synthetic and real-world datasets.

\italicparagraph{Replica.}
Following GSN~\cite{devries2021gsn}, we use Replica~\cite{replica} for evaluation.
The dataset contains 18 realistic synthetic scenes, and we generate the ground truth by the Habitat~\cite{habitat2019,habitat2021} renderer.
Specifically, we first divide the 18 scenes into 48 disjoint sub-areas with similar space sizes, and randomly sample 300 camera poses that fit the human perspective for each sub-area.
Then we collect the RGB and depth observations in 256$\times$256 resolution for each camera pose. 
We finally select 15 scenes (42 sub-areas) for training and the rest 3 scenes (6 sub-areas) for validation.
The generation is performed on a triplet of two input source views and one output query view. 
For each query view, the two source views are selected by satisfying: (1) no overlap; (2) less than 50\% overlap with the query view; and (3) totally 65\%-70\% overlap with the query view, which means that the 30\%-35\% of the pixels in the query view need inpainting.
For each sampled camera pose, we randomly select 3 triplets meeting the requirements, which results in $\sim$37k triplets for training and $\sim$5k for validation.

\italicparagraph{ARKitScenes.}
The real-world dataset, ARKitScenes~\cite{dehghan2021arkitscenes}, is one of the largest datasets for indoor scene understanding.
The dataset consists of 5,047 captures of 1,661 unique scenes, with high-quality ground truth of registered RGB-D frames.
Due to its huge amount of data, for each scene, we only select its longest capture and uniformly sample it with an interval of 10 frames ($\sim$1s).
We adopt the same strategy described above in the Replica dataset to generate triplets but adjust the overlap rates of the query view with each individual source view and two total source views to 60\% and 60\%-80\% respectively.
Considering the temporal relationship between frames, the suitable source views are searched within a window of 9 frames centered by the query view for efficiency.
We follow ARKitScenes' official train-val split and finally obtain $\sim$15k triplets for training and $\sim$2k triplets for validation.
The image resolution is 256$\times$192.

\boldparagraph{Implementation details.}
Our framework is implemented in PyTorch and run on a single NVIDIA Tesla V100 GPU with 32GB memory.
The voxel size of all the scenes in our experiment is chosen to be 10cm, and the embedding size is set as $d=32$, (8 for alpha value and 24 for RGB).
The point cloud encoder, geometry predictor, and texture inpainter are all implemented via the Minkowski Engine~\cite{choy2019minkowski} and adopt their provided default network architectures with sparse voxel convolutions such as ResNet34, generative U-Net and the vanilla U-Net.
The training takes around 6 days in total to go through all the training steps from scratch.
The experiments are conducted on the validation sets of Replica and ARKitScenes.
We not only compare the results of the target view, but also the short video generated based on the poses between the target view and the two source views.
For Replica, the intermediate poses are linearly interpolated, and the ground-truth images are generated based on these poses using Habitat.
For ARKitScenes, we uniformly sample intermediate poses from the original scan sequence between the source views and the target view.

\boldparagraph{Metrics.}
For quantitative evaluation, we mainly follow pixelNeRF~\cite{yu2021pixelnerf} and PixelSynth~\cite{rockwell2021pixelsynth} to use SSIM as low-level metrics to measure the window differences between the ground truth and the prediction.
The high-level measures such as FID~\cite{heusel2017fid} and LPIPS~\cite{zhang2018perceptual} (perceptual similarity) are also taken into account.
For the commonly used PSNR, as explained also in PixelSynth~\cite{rockwell2021pixelsynth} paper, it may not be a good metric in such an extrapolation task. 
Therefore, we report these numbers here for reference only.
We also compare the generated geometry separately.
Since the scene representations of each method are different, for convenience, we directly compare the rendered depth of the center frame by RMSE and $\delta^{1.25}$, which are commonly used in the field of depth estimation.
The purpose of the video result comparison is to better evaluate the temporal consistency of the textures across frames, so we only report the image quality metrics for videos.
The FID score of videos is also excluded because it uses globally pooled features for calculation, where both the temporal and spatial relationships are lost.
Additionally, we incorporated a user study of 30 people to evaluate the quality of the generated videos, which consists of 20 multiple-choice questions (half from Replica and half from ARKitScenes). 
For each question, the participant was asked to select a single generated video from four options (baselines and ours) that they considered more visually plausible.

\subsection{Baseline Comparison}

\boldparagraph{Baselines.}
We select PixelSynth~\cite{rockwell2021pixelsynth}, pixelNeRF~\cite{yu2021pixelnerf} and SPSG~\cite{dai2021spsg} as baselines.
Since they are not original proposed for our task and the depths of the input views are required in the task, we make necessary adaptions on these baselines for a fair comparison, which are briefly described in the following.

\italicparagraph{PixelSynth}~\cite{rockwell2021pixelsynth} outpaints the missing parts in a chosen support view, which is then back-projected to the 3D space followed by the point cloud rendering technology with refinement.
Here our target view is used as its support view.
The depth of the input view in the original paper is predicted by a network, but in our task we directly provide the depth observations as input.
The depth estimation module still exists in the adapted version for predicting the depth of the generated part.
The rendering and refinement procedures follows the original.

\italicparagraph{pixelNeRF}~\cite{yu2021pixelnerf} does not require depth inputs. 
The feature acquisition is done by projecting the sampling points on the rays of the target view to the source views and extracting encoded features of the corresponding pixel locations. 
Here we utilize the depths of the source views and back-project the features to the 3D space, where the $k$-NN search is afterward performed to more precisely obtain the features for the sampling points. 
The following procedures such as ray feature aggregation and rendering are the same as the original. 
Due to the additional searching phase, the runtime of this adapted version is slightly longer.

\italicparagraph{SPSG}~\cite{dai2021spsg} generates 3D models with both geometry and color from incomplete RGB-D scans, and raycasting is applied to render the target frames in specific poses. 
We turn the source views into the TSDF through volumetric fusion as the required input for SPSG, followed by the same original procedure.
SPSG avoids memory issues by tessellating the scene into smaller chunks (64$\times$64$\times$128) for training, however, with a small voxel size (\eg, 2cm), the unobserved area is typically larger than such a chuck, thus unable to generate the desired data to train the baseline.
To avoid the problem, the voxel scale is still similarly set to 10cm to be consistent with our approach, and the scene is processed and completed as a whole for a fair comparison.

\begin{table*}[!t]
    \centering
    \scriptsize
    \setlength{\tabcolsep}{2.8pt}
    \renewcommand{\arraystretch}{1.2}
    \newcommand{\csp}{\hskip 0.27em}
    
    \caption{\textbf{Quantitative baseline comparison on the generated query view for Replica~\cite{replica} dataset}. Numbers of each entry: full frame / unobserved regions.}
	\begin{tabular}{@{}l@{\csp}|@{\csp}cc@{\csp}|@{\csp}cc@{\csp}|@{\csp}cc@{}}
	\toprule
	\textbf{Method} & PSNR$\uparrow$ & SSIM$\uparrow$ & FID$\downarrow$ & LPIPS$\downarrow$ & Dep. RMSE$\downarrow$ & Dep. $\delta^{1.25}\uparrow$ \\
    \midrule
	SPSG~\cite{dai2021spsg} & 29.12/29.36 & 0.603/0.452 & 175.2/207.6 & 0.518/0.804 & 0.914/1.243 & 0.790/0.534 \\
	pixelNeRF~\cite{yu2021pixelnerf} & 28.78/28.90 & 0.650/0.583 & 257.3/215.8 & 0.610/0.592 & 0.713/0.793 & 0.560/0.491 \\
	PixelSynth~\cite{rockwell2021pixelsynth} & 31.28/30.44 & 0.633/0.587 & 80.58/\best{177.4} & \best{0.315}/0.576 & \best{0.382}/0.534 & 0.869/0.664  \\
    \midrule
    
	Ours & 29.65/29.42 & \best{0.706}/\best{0.600} & \best{74.84}/182.6 & 0.332/\best{0.575} & 0.397/\best{0.456} & \best{0.879}/\best{0.817} \\
	\bottomrule
    \end{tabular}
	\label{tab:replica}
\end{table*}
\begin{table*}[!t]
    \centering
    \scriptsize
    \setlength{\tabcolsep}{2pt}
    \renewcommand{\arraystretch}{1.2}
    \newcommand{\csp}{\hskip 0.16em}
    \caption{\textbf{Quantitative baseline comparison on the generated query view for ARKitScenes~\cite{dehghan2021arkitscenes} dataset}. Numbers of each entry: full frame / unobserved regions.}
	\begin{tabular}{@{}l@{\csp}|@{\csp}cc@{\csp}|@{\csp}cc@{\csp}|@{\csp}cc@{}}
	\toprule
	\textbf{Method} & PSNR$\uparrow$ & SSIM$\uparrow$ & FID$\downarrow$ & LPIPS$\downarrow$ & Dep. RMSE$\downarrow$ & Dep. $\delta^{1.25}\uparrow$ \\

    \midrule
    
	SPSG~\cite{dai2021spsg} & 28.41/28.34 & 0.406/0.293 & 175.43/188.00 & 0.522/0.875 & 0.680/0.918 & 0.763/0.661 \\
	
	pixelNeRF~\cite{yu2021pixelnerf} & 28.12/28.10 & 0.504/\best{0.442} & 262.65/189.01 & 0.724/1.065 & 0.546/0.541 & 0.589/0.553 \\
	PixelSynth~\cite{rockwell2021pixelsynth} & 28.42/28.40 & 0.385/0.260 & \best{101.96}/\best{149.72} & 0.496/0.920 & 0.445/0.696 & 0.779/0.397 \\
    \midrule
    
	Ours & 28.38/28.10 & \best{0.551}/0.414 & 141.40/160.57 & \best{0.427}/\best{0.733} & \best{0.394}/\best{0.489} & \best{0.927}/\best{0.883} \\
	\bottomrule
    \end{tabular}
	\label{tab:arkit}
\end{table*}
\begin{table}[!t]
    \centering
    \scriptsize
    \setlength{\tabcolsep}{4pt}
    \renewcommand{\arraystretch}{1.2}
    \newcommand{\csp}{\hskip 0.7em}
    \caption{\textbf{Quantitative baseline comparison on the generated videos}. }
    \begin{tabular}{l@{\csp}|@{\csp}cccc@{\csp}|@{\csp}cccc@{\csp}}
    \toprule
    & \multicolumn{4}{c@{\csp}|@{\csp}}{\textbf{Replica}\cite{replica}} & \multicolumn{4}{c}{\textbf{ARKitScenes}\cite{dehghan2021arkitscenes}} \\
    \textbf{Method} & PSNR$\uparrow$ & SSIM$\uparrow$ & LPIPS$\downarrow$ & Human$\uparrow$ & PSNR$\uparrow$ & SSIM$\uparrow$ & LPIPS$\downarrow$ & Human$\uparrow$ \\
    \midrule
    SPSG~\cite{dai2021spsg} & 27.91 & 0.521 & 0.694 & 1.3\% & 28.27 & 0.427 & 0.518 & 2.3\% \\
    pixelNeRF~\cite{yu2021pixelnerf} & 28.95 & 0.670 & 0.624 & 1.7\% & 28.96 & 0.524 & 0.727 & 0.0\% \\
    PixelSynth~\cite{rockwell2021pixelsynth} & 31.42 & 0.605 & 0.323 & 16.7\% & 28.95 & 0.393 & 0.463 & 43.7\% \\
    \midrule
    \textbf{Ours} & 29.83 & \best{0.728} & \best{0.301} & \best{80.3\%} & 28.66 & \best{0.589} & \best{0.402} & \best{54.0\%} \\
    \bottomrule
    \end{tabular}
    \label{tab:video}
\end{table}

\begin{figure*}[!ht]
    \centering
    \scriptsize
    \newcommand{\sz}{0.16\linewidth} 

    \noindent\animategraphics[autoplay,loop,height=\sz,palindrome,poster=\POSTERFRAME]{5}{images/Replica_Input/13_035_045_104/}{\STARTFRAME}{\ENDFRAME}
    \noindent\animategraphics[autoplay,loop,height=\sz,palindrome,poster=\POSTERFRAME]{5}{images/Replica_GT/13_035_045_104/}{\STARTFRAME}{\ENDFRAME}
    \noindent\animategraphics[autoplay,loop,height=\sz,palindrome,poster=\POSTERFRAME]{5}{images/Replica_SPSG/13_035_045_104/}{\STARTFRAME}{\ENDFRAME}
    \noindent\animategraphics[autoplay,loop,height=\sz,palindrome,poster=\POSTERFRAME]{5}{images/Replica_PixelNerf/13_035_045_104/}{\STARTFRAME}{\ENDFRAME}
    \noindent\animategraphics[autoplay,loop,height=\sz,palindrome,poster=\POSTERFRAME]{5}{images/Replica_PixelSynth/13_035_045_104/}{\STARTFRAME}{\ENDFRAME}
    \noindent\animategraphics[autoplay,loop,height=\sz,palindrome,poster=\POSTERFRAME]{5}{images/Replica_Ours/13_035_045_104/}{\STARTFRAME}{\ENDFRAME}
    \\ [1pt]
    \noindent\animategraphics[autoplay,loop,height=\sz,palindrome,poster=\POSTERFRAME]{5}{images/Replica_Input/13_004_105_253/}{\STARTFRAME}{\ENDFRAME}
    \noindent\animategraphics[autoplay,loop,height=\sz,palindrome,poster=\POSTERFRAME]{5}{images/Replica_GT/13_004_105_253/}{\STARTFRAME}{\ENDFRAME}
    \noindent\animategraphics[autoplay,loop,height=\sz,palindrome,poster=\POSTERFRAME]{5}{images/Replica_SPSG/13_004_105_253/}{\STARTFRAME}{\ENDFRAME}
    \noindent\animategraphics[autoplay,loop,height=\sz,palindrome,poster=\POSTERFRAME]{5}{images/Replica_PixelNerf/13_004_105_253/}{\STARTFRAME}{\ENDFRAME}
    \noindent\animategraphics[autoplay,loop,height=\sz,palindrome,poster=\POSTERFRAME]{5}{images/Replica_PixelSynth/13_004_105_253/}{\STARTFRAME}{\ENDFRAME}
    \noindent\animategraphics[autoplay,loop,height=\sz,palindrome,poster=\POSTERFRAME]{5}{images/Replica_Ours/13_004_105_253/}{\STARTFRAME}{\ENDFRAME}
    \\ [1pt]
    \noindent\animategraphics[autoplay,loop,height=\sz,palindrome,poster=\POSTERFRAME]{5}{images/Replica_Input/14_183_114_278/}{\STARTFRAME}{\ENDFRAME}
    \noindent\animategraphics[autoplay,loop,height=\sz,palindrome,poster=\POSTERFRAME]{5}{images/Replica_GT/14_183_114_278/}{\STARTFRAME}{\ENDFRAME}
    \noindent\animategraphics[autoplay,loop,height=\sz,palindrome,poster=\POSTERFRAME]{5}{images/Replica_SPSG/14_183_114_278/}{\STARTFRAME}{\ENDFRAME}
    \noindent\animategraphics[autoplay,loop,height=\sz,palindrome,poster=\POSTERFRAME]{5}{images/Replica_PixelNerf/14_183_114_278/}{\STARTFRAME}{\ENDFRAME}
    \noindent\animategraphics[autoplay,loop,height=\sz,palindrome,poster=\POSTERFRAME]{5}{images/Replica_PixelSynth/14_183_114_278/}{\STARTFRAME}{\ENDFRAME}
    \noindent\animategraphics[autoplay,loop,height=\sz,palindrome,poster=\POSTERFRAME]{5}{images/Replica_Ours/14_183_114_278/}{\STARTFRAME}{\ENDFRAME}
    \\ [1pt]
    \noindent\animategraphics[autoplay,loop,height=\sz,palindrome,poster=\POSTERFRAME]{5}{images/Replica_Input/19_116_273_282/}{\STARTFRAME}{\ENDFRAME}
    \noindent\animategraphics[autoplay,loop,height=\sz,palindrome,poster=\POSTERFRAME]{5}{images/Replica_GT/19_116_273_282/}{\STARTFRAME}{\ENDFRAME}
    \noindent\animategraphics[autoplay,loop,height=\sz,palindrome,poster=\POSTERFRAME]{5}{images/Replica_SPSG/19_116_273_282/}{\STARTFRAME}{\ENDFRAME}
    \noindent\animategraphics[autoplay,loop,height=\sz,palindrome,poster=\POSTERFRAME]{5}{images/Replica_PixelNerf/19_116_273_282/}{\STARTFRAME}{\ENDFRAME}
    \noindent\animategraphics[autoplay,loop,height=\sz,palindrome,poster=\POSTERFRAME]{5}{images/Replica_PixelSynth/19_116_273_282/}{\STARTFRAME}{\ENDFRAME}
    \noindent\animategraphics[autoplay,loop,height=\sz,palindrome,poster=\POSTERFRAME]{5}{images/Replica_Ours/19_116_273_282/}{\STARTFRAME}{\ENDFRAME}
    \\ [1pt]
    \noindent\animategraphics[autoplay,loop,height=\sz,palindrome,poster=\POSTERFRAME]{5}{images/Replica_Input/20_183_120_192/}{\STARTFRAME}{\ENDFRAME}
    \noindent\animategraphics[autoplay,loop,height=\sz,palindrome,poster=\POSTERFRAME]{5}{images/Replica_GT/20_183_120_192/}{\STARTFRAME}{\ENDFRAME}
    \noindent\animategraphics[autoplay,loop,height=\sz,palindrome,poster=\POSTERFRAME]{5}{images/Replica_SPSG/20_183_120_192/}{\STARTFRAME}{\ENDFRAME}
    \noindent\animategraphics[autoplay,loop,height=\sz,palindrome,poster=\POSTERFRAME]{5}{images/Replica_PixelNerf/20_183_120_192/}{\STARTFRAME}{\ENDFRAME}
    \noindent\animategraphics[autoplay,loop,height=\sz,palindrome,poster=\POSTERFRAME]{5}{images/Replica_PixelSynth/20_183_120_192/}{\STARTFRAME}{\ENDFRAME}
    \noindent\animategraphics[autoplay,loop,height=\sz,palindrome,poster=\POSTERFRAME]{5}{images/Replica_Ours/20_183_120_192/}{\STARTFRAME}{\ENDFRAME}
    \\ [1pt]
    \noindent\animategraphics[autoplay,loop,height=\sz,palindrome,poster=\POSTERFRAME]{5}{images/Replica_Input/21_027_002_152/}{\STARTFRAME}{\ENDFRAME}
    \noindent\animategraphics[autoplay,loop,height=\sz,palindrome,poster=\POSTERFRAME]{5}{images/Replica_GT/21_027_002_152/}{\STARTFRAME}{\ENDFRAME}
    \noindent\animategraphics[autoplay,loop,height=\sz,palindrome,poster=\POSTERFRAME]{5}{images/Replica_SPSG/21_027_002_152/}{\STARTFRAME}{\ENDFRAME}
    \noindent\animategraphics[autoplay,loop,height=\sz,palindrome,poster=\POSTERFRAME]{5}{images/Replica_PixelNerf/21_027_002_152/}{\STARTFRAME}{\ENDFRAME}
    \noindent\animategraphics[autoplay,loop,height=\sz,palindrome,poster=\POSTERFRAME]{5}{images/Replica_PixelSynth/21_027_002_152/}{\STARTFRAME}{\ENDFRAME}
    \noindent\animategraphics[autoplay,loop,height=\sz,palindrome,poster=\POSTERFRAME]{5}{images/Replica_Ours/21_027_002_152/}{\STARTFRAME}{\ENDFRAME}
    \\
    
    \setlength{\tabcolsep}{1pt}
    \begin{tabular}{C{\sz}C{\sz}C{\sz}C{\sz}C{\sz}C{\sz}}
    Input & GT & SPSG~\cite{dai2021spsg} & pixelNeRF~\cite{yu2021pixelnerf} & PixelSynth~\cite{rockwell2021pixelsynth} & \textbf{Ours}
    \end{tabular}
    
    \caption{\textbf{Qualitative baseline comparison on the Replica~\cite{replica} dataset.} We show comparisons to the state of the art on a variety of examples. Our method generates more realistic videos with better temporal consistency and fewer artifacts. \USEADOBE}
    \label{fig:replica}

\end{figure*}

\begin{figure*}[!ht]
    \centering
    \scriptsize
    \newcommand{\sz}{0.16\linewidth} 

    \noindent\animategraphics[autoplay,loop,width=\sz,palindrome,poster=\POSTERFRAME]{5}{images/Arkit_Input/41159519_38253007_38252007_38256006/}{\STARTFRAME}{\ENDFRAME}
    \noindent\animategraphics[autoplay,loop,width=\sz,palindrome,poster=\POSTERFRAME]{5}{images/Arkit_GT/41159519_38253007_38252007_38256006/}{\STARTFRAME}{\ENDFRAME}
    \noindent\animategraphics[autoplay,loop,width=\sz,palindrome,poster=\POSTERFRAME]{5}{images/Arkit_SPSG/41159519_38253007_38252007_38256006/}{\STARTFRAME}{\ENDFRAME}
    \noindent\animategraphics[autoplay,loop,width=\sz,palindrome,poster=\POSTERFRAME]{5}{images/Arkit_PixelNerf/41159519_38253007_38252007_38256006/}{\STARTFRAME}{\ENDFRAME}
    \noindent\animategraphics[autoplay,loop,width=\sz,palindrome,poster=\POSTERFRAME]{5}{images/Arkit_PixelSynth/41159519_38253007_38252007_38256006/}{\STARTFRAME}{\ENDFRAME}
    \noindent\animategraphics[autoplay,loop,width=\sz,palindrome,poster=\POSTERFRAME]{5}{images/Arkit_Ours/41159519_38253007_38252007_38256006/}{\STARTFRAME}{\ENDFRAME}
    \\ [1pt]
    \noindent\animategraphics[autoplay,loop,width=\sz,palindrome,poster=\POSTERFRAME]{5}{images/Arkit_Input/41159519_38305002_38303003_38307001/}{\STARTFRAME}{\ENDFRAME}
    \noindent\animategraphics[autoplay,loop,width=\sz,palindrome,poster=\POSTERFRAME]{5}{images/Arkit_GT/41159519_38305002_38303003_38307001/}{\STARTFRAME}{\ENDFRAME}
    \noindent\animategraphics[autoplay,loop,width=\sz,palindrome,poster=\POSTERFRAME]{5}{images/Arkit_SPSG/41159519_38305002_38303003_38307001/}{\STARTFRAME}{\ENDFRAME}
    \noindent\animategraphics[autoplay,loop,width=\sz,palindrome,poster=\POSTERFRAME]{5}{images/Arkit_PixelNerf/41159519_38305002_38303003_38307001/}{\STARTFRAME}{\ENDFRAME}
    \noindent\animategraphics[autoplay,loop,width=\sz,palindrome,poster=\POSTERFRAME]{5}{images/Arkit_PixelSynth/41159519_38305002_38303003_38307001/}{\STARTFRAME}{\ENDFRAME}
    \noindent\animategraphics[autoplay,loop,width=\sz,palindrome,poster=\POSTERFRAME]{5}{images/Arkit_Ours/41159519_38305002_38303003_38307001/}{\STARTFRAME}{\ENDFRAME}
    \\ [1pt]
    \noindent\animategraphics[autoplay,loop,width=\sz,palindrome,poster=\POSTERFRAME]{5}{images/Arkit_Input/41159566_35355614_35354614_35358613/}{\STARTFRAME}{\ENDFRAME}
    \noindent\animategraphics[autoplay,loop,width=\sz,palindrome,poster=\POSTERFRAME]{5}{images/Arkit_GT/41159566_35355614_35354614_35358613/}{\STARTFRAME}{\ENDFRAME}
    \noindent\animategraphics[autoplay,loop,width=\sz,palindrome,poster=\POSTERFRAME]{5}{images/Arkit_SPSG/41159566_35355614_35354614_35358613/}{\STARTFRAME}{\ENDFRAME}
    \noindent\animategraphics[autoplay,loop,width=\sz,palindrome,poster=\POSTERFRAME]{5}{images/Arkit_PixelNerf/41159566_35355614_35354614_35358613/}{\STARTFRAME}{\ENDFRAME}
    \noindent\animategraphics[autoplay,loop,width=\sz,palindrome,poster=\POSTERFRAME]{5}{images/Arkit_PixelSynth/41159566_35355614_35354614_35358613/}{\STARTFRAME}{\ENDFRAME}
    \noindent\animategraphics[autoplay,loop,width=\sz,palindrome,poster=\POSTERFRAME]{5}{images/Arkit_Ours/41159566_35355614_35354614_35358613/}{\STARTFRAME}{\ENDFRAME}
    \\ [1pt]
    \noindent\animategraphics[autoplay,loop,width=\sz,palindrome,poster=\POSTERFRAME]{5}{images/Arkit_Input/42898497_749915075_749914076_749916075/}{\STARTFRAME}{\ENDFRAME}
    \noindent\animategraphics[autoplay,loop,width=\sz,palindrome,poster=\POSTERFRAME]{5}{images/Arkit_GT/42898497_749915075_749914076_749916075/}{\STARTFRAME}{\ENDFRAME}
    \noindent\animategraphics[autoplay,loop,width=\sz,palindrome,poster=\POSTERFRAME]{5}{images/Arkit_SPSG/42898497_749915075_749914076_749916075/}{\STARTFRAME}{\ENDFRAME}
    \noindent\animategraphics[autoplay,loop,width=\sz,palindrome,poster=\POSTERFRAME]{5}{images/Arkit_PixelNerf/42898497_749915075_749914076_749916075/}{\STARTFRAME}{\ENDFRAME}
    \noindent\animategraphics[autoplay,loop,width=\sz,palindrome,poster=\POSTERFRAME]{5}{images/Arkit_PixelSynth/42898497_749915075_749914076_749916075/}{\STARTFRAME}{\ENDFRAME}
    \noindent\animategraphics[autoplay,loop,width=\sz,palindrome,poster=\POSTERFRAME]{5}{images/Arkit_Ours/42898497_749915075_749914076_749916075/}{\STARTFRAME}{\ENDFRAME}
    \\ [1pt]
    \noindent\animategraphics[autoplay,loop,width=\sz,palindrome,poster=\POSTERFRAME]{5}{images/Arkit_Input/45663099_54885049_54884049_54886048/}{\STARTFRAME}{\ENDFRAME}
    \noindent\animategraphics[autoplay,loop,width=\sz,palindrome,poster=\POSTERFRAME]{5}{images/Arkit_GT/45663099_54885049_54884049_54886048/}{\STARTFRAME}{\ENDFRAME}
    \noindent\animategraphics[autoplay,loop,width=\sz,palindrome,poster=\POSTERFRAME]{5}{images/Arkit_SPSG/45663099_54885049_54884049_54886048/}{\STARTFRAME}{\ENDFRAME}
    \noindent\animategraphics[autoplay,loop,width=\sz,palindrome,poster=\POSTERFRAME]{5}{images/Arkit_PixelNerf/45663099_54885049_54884049_54886048/}{\STARTFRAME}{\ENDFRAME}
    \noindent\animategraphics[autoplay,loop,width=\sz,palindrome,poster=\POSTERFRAME]{5}{images/Arkit_PixelSynth/45663099_54885049_54884049_54886048/}{\STARTFRAME}{\ENDFRAME}
    \noindent\animategraphics[autoplay,loop,width=\sz,palindrome,poster=\POSTERFRAME]{5}{images/Arkit_Ours/45663099_54885049_54884049_54886048/}{\STARTFRAME}{\ENDFRAME}
    \\
    
    \setlength{\tabcolsep}{1pt}
    \begin{tabular}{C{\sz}C{\sz}C{\sz}C{\sz}C{\sz}C{\sz}}
    Input & GT & SPSG~\cite{dai2021spsg} & PixelNeRF~\cite{yu2021pixelnerf} & PixelSynth~\cite{rockwell2021pixelsynth} & \textbf{Ours}
    \end{tabular}
    
    \caption{\textbf{Qualitative baseline comparison on the ARKitScenes~\cite{dehghan2021arkitscenes} dataset.} We show comparisons to the state of the art on a variety of examples. Our method generates more realistic videos with better temporal consistency and fewer artifacts. \USEADOBE}
    \label{fig:arkit}

\end{figure*}

\boldparagraph{Quantitative results.}
We show quantitative baseline comparisons for the \textbf{\textit{center frame (query view)}} on Replica and ARKitScenes in \cref{tab:replica} and \cref{tab:arkit}, while the quantitative results in terms of \textbf{\textit{videos}} are reported in \cref{tab:video}.

\italicparagraph{Texture.}
On the low-level metric SSIM, our method outperforms all other baselines on both datasets and all the area levels (the unobserved part, the whole query view, and the video) except the unobserved part of the ARKitScenes dataset.
This may indicate that although pixelNeRF generates blurry results in \cref{fig:arkit}, the pixel color distribution of the local windows in the unobserved part could still be close to GT.
In terms of perceptual similarity, our method outperforms all other baselines on both datasets and all the area levels except the full query view of the Replica dataset.
Since PixelSynth's LPIPS score is similar to ours on the generated part, it could be inferred that PixelSynth may better perform on the observed part on the synthetic dataset.
This may be because, in the case of a synthetic dataset, the geometry is rather simple and does not have many occlusions. Details could be recovered better through the point cloud projection they use in this case, rather than a voxel representation.
For the FID score, ours performs on par with PixelSynth on the Replica dataset, but slightly worse on ARKitScenes, which means that PixelSynth may have a higher capacity to produce richer textures in the real-world dataset.
However, since FID only represents the distribution between the overall features of the two sets of images, it generally cannot measure the spatial similarity of two images.

\italicparagraph{Geometry.}
In terms of geometric similarity, measured by depth RMSE and $\delta^{1.25}$, our method can predict more accurate depth in most cases.
This could be the main reason that leads to a better 3D consistency of our generated videos and better performance in high-level metrics of videos in \cref{tab:ablation}.
On the other hand, the depths in the generated part of all methods are almost worse than that of the whole image, however, the drop of PixelSynth is relatively large. 
It can be inferred that the geometry predicted by PixelSynth of the unobserved part does not match the input very well. 
This is also reflected in the video results in \cref{fig:replica} and \cref{fig:arkit}, where its generated parts are disjointed from time to time.

\italicparagraph{User study.}
We list the user study results in \cref{tab:video} (``Human'' column). 
Compared to the baselines, we found that ours achieved higher average approval rates for the generated video quality. 
On the ARKitScenes dataset, PixelSynth scores second place with a much smaller gap. %
We believe this is mainly due to that the geometry generated by PixelSynth on ARKitScenes is generally better than that of Replica, and the gap between the generated part and the observation is much smaller, leading to a better 3D-consistency in the synthesized video.

\boldparagraph{Qualitative results.}
\cref{fig:replica} and \cref{fig:arkit} visualize the generated videos on selected samples of the Replica and ARKitScenes datasets.
It can be seen that the completion results by SPSG still have many empty spaces. 
We believe this is because SPSG is originally proposed for filling small holes on denser input and has difficulties adapting to our sparser input with large gaps.
pixelNeRF's results are much more blurry than other columns.
We attribute this to the low resolutions of features used by pixelNeRF for image rendering.
PixelSynth is the most competitive baseline method for predicting relatively complete and sharp results. 
However, it tends to generate results that are less plausible in the visualization. 
For instance, on the second sample of \cref{fig:replica}, its prediction on the corner resembles the entrance of a new room, which has low geometrical plausibility with the neighboring observed walls. 
We believe that the absence of geometry information in the texture completion is the main reason for the implausible results of PixelSynth. 
Compared with the baseline methods, our method can better fill the gaps in scenes and produces more plausible outputs for both texture and geometry. 
For example, on the third sample of \cref{fig:arkit} with complicated texture, our method fills the gap with a ladder and bookshelf which are consistent with the contextual texture. 
On the final sample of \cref{fig:arkit} with high variance in the spatial structure, our method predicts better geometry of the wall and roof. 
However, compared with PixelSynth, our method comes short of the sharpness in the generated frames. 
This is mainly limited by the voxel embeddings used in our pipeline which is bound by the voxel resolution (10cm$^3$ size). 
Also, on many validation triplets, there are objects in the query view that are totally unobservable in the source views, \eg, the door in the second sample of \cref{fig:replica} and the shelf in the fourth sample of \cref{fig:arkit}. 
In these empty regions, both methods tend to fill contents that are different from the ground truth. 
Hence, neither method can achieve an excellent score on the high-level metrics.

\subsection{Ablation Study}

\begin{table*}[!t]
    \centering
    \scriptsize
    \setlength{\tabcolsep}{9pt}
    \newcommand{\onezero}{\textcolor{white}{0}}
    \newcommand{\twozero}{\textcolor{white}{00}}
    \newcommand{\threezero}{\textcolor{white}{000}}
	\newcommand{\XXdotXX}{\twozero-\twozero}
	\newcommand{\XdotXXX}{\onezero-\threezero}
    \renewcommand{\arraystretch}{1.2}
    \newcommand{\csp}{\hskip 1.8em}
    
    \caption[]{\textbf{Quantitative ablation study on the generated query view.} The ablations are: \textbf{B}: basic framework; \textbf{+E}: point cloud encoder; \textbf{+D}: disentangled MLPs; \textbf{+R}: refinement module. Numbers of each entry: full frame / \textbf{\textit{unobserved}} regions.}

	\begin{tabular}{@{}l@{\csp}|@{\csp}cc@{\csp}|@{\csp}cccc@{}}
	\toprule
	\textbf{Method} & PSNR$\uparrow$ & SSIM$\uparrow$ & FID$\downarrow$ & LPIPS$\downarrow$ \\


    \midrule
    B & 30.06/29.38 & 0.730/0.636 & 128.5/210.2 & 0.418/0.613 \\
    B+E & 29.67/29.39 & \best{0.746}/\best{0.644} & 135.0/210.7 & 0.395/0.628 \\
	B+E+D & 29.58/29.37 & 0.729/0.603 & 124.2/216.3 & 0.404/0.672 \\
    \midrule
    B+E+D+R (Ours) & 29.65/29.42 & 0.706/0.600 & \best{74.84}/\best{182.6} & \best{0.332}/\best{0.575} \\
	\bottomrule
    \end{tabular}

	\label{tab:ablation}
\end{table*}

\begin{figure*}[!t]
    \centering
    \scriptsize
    \newcommand{\sz}{0.16\linewidth} 

    \includegraphics[width=\sz]{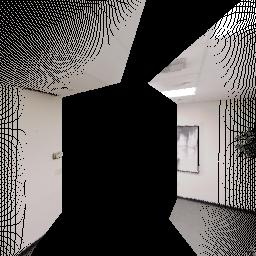}
    \includegraphics[width=\sz]{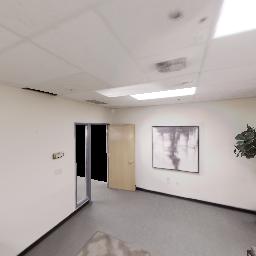}
    \includegraphics[width=\sz]{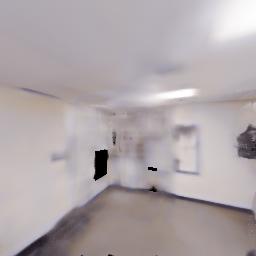}
    \includegraphics[width=\sz]{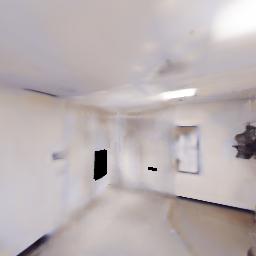}
    \includegraphics[width=\sz]{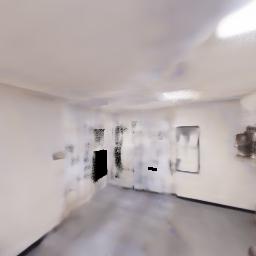}
    \includegraphics[width=\sz]{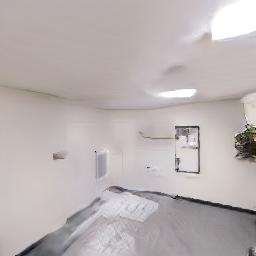}
    \\ [1pt]
    \includegraphics[width=\sz]{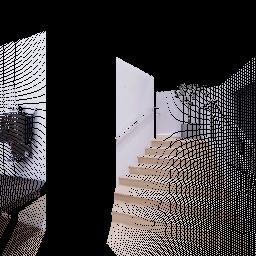}
    \includegraphics[width=\sz]{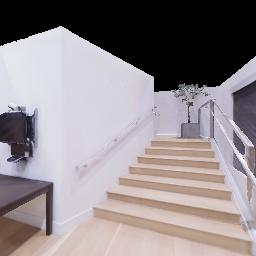}
    \includegraphics[width=\sz]{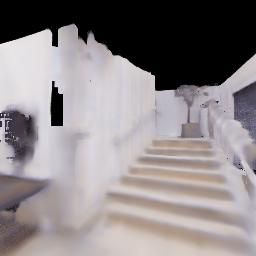}
    \includegraphics[width=\sz]{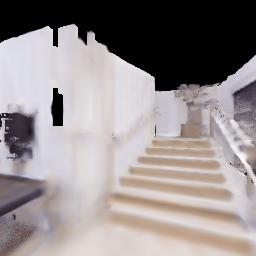}
    \includegraphics[width=\sz]{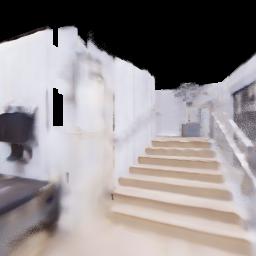}
    \includegraphics[width=\sz]{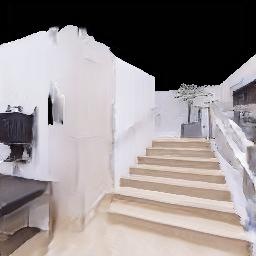}
    \\ [1pt]
    \includegraphics[width=\sz]{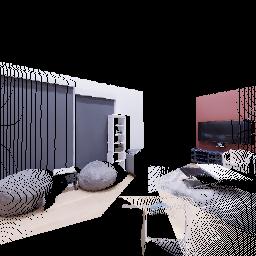}
    \includegraphics[width=\sz]{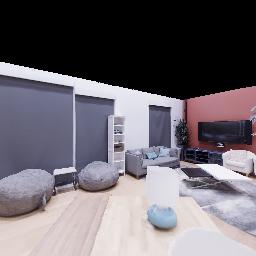}
    \includegraphics[width=\sz]{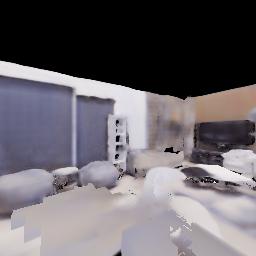}
    \includegraphics[width=\sz]{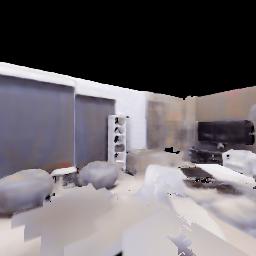}
    \includegraphics[width=\sz]{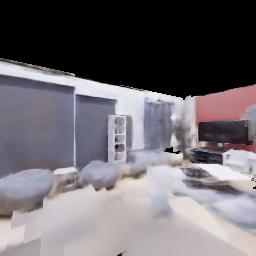}
    \includegraphics[width=\sz]{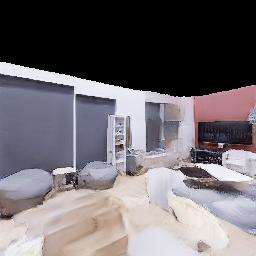}
    \\
    
    \setlength{\tabcolsep}{1pt}
    \begin{tabular}{C{\sz}C{\sz}C{\sz}C{\sz}C{\sz}C{\sz}}
    Input & GT & B & B+E & B+E+D & B+E+D+R
    \end{tabular}
    
    \caption{\textbf{Qualitative ablation study on the generated query view.} We present exemplary qualitative results for various ablations of our method.
	The rendered videos visibly contain more details and achieve higher levels realism with our full method.}
    \label{fig:abl}
\end{figure*}

To better evaluate the effectiveness of the individual components of our method, we also conduct an ablation study on Replica~\cite{replica} by incrementally adding components to our basic framework \textbf{(B)}.
More specifically, we focus on the following three components: \textbf{(E)} the encoder used for processing the point cloud input; \textbf{(D)} the disentangled MLPs for RGB and alpha value; and \textbf{(R)} the 2D refinement module.
In the basic framework, we directly voxelize the point cloud with RGB according to the defined voxel size as the input of the 3D generation pipeline, which outputs completed embeddings.
All the ablations use the same geometry completion network which does not take color information as input.

\cref{tab:ablation} shows quantitative evaluation results of the ablation study and \cref{fig:abl} illustrates the center frame generated by different ablative pipelines.
It can be observed that frames generated by the full pipeline show the best performance in high-level measurements (FID and LPIPS).
Comparing between results of B and B+E, we find that encoding the point cloud into the voxel embeddings via an encoder facilitates the texture generation and preserves more details, which is also consistent with the SSIM increase and LPIPS decrease.
We still found a deviation between the generated colors and the GT.
We believe this is due to the entangled embeddings having to take into account both geometry and texture information.
After integrating D, it can be seen that the recovered colors are further improved and closer to the ground truth. 
For example, the pink wall in the third sample of \cref{fig:abl} is recovered to be pink after adding D. 
This indicates the disentangled prediction of alpha values and RGB values can improve the modeling of color patterns in scenes.
Finally, the 2D refinement module (R) increases the details in the results as well as fills the small holes caused by the geometric incompleteness.
One notable point shown in the table is that the SSIM decreases after adding disentangled MLPs (D) and the refinement module (R) even though they can produce sharper image results than B and B+E. 
Based on our experiments, we find that SSIM scores drop less when the image is blurred while the general structure is maintained. 
Likewise, sharpening an image does not improve SSIM scores much.
We believe the pipeline scarifies structural similarity scores in favor of improving the visual quality.

\subsection{Limitations}
%

Although our method can efficiently complete a scene, generate a high-quality video from novel views, and outperform existing methods, there are three major limitations:
(1) The completion of geometry can sometimes produce holes and incomplete structures, which result in visible artifacts.
(2) The introduction of the 2D upsampling step may slightly break the strict 3D consistency obtained from rendering and frame-to-frame flickering is possible for individual pixels though this is rather unlikely due to local context consistency.
(3) The voxel resolution and thus the voxel size is limited in our pipeline. 
A too-large voxel size may blur the result while a too-small value could lead to a huge number of voxels resulting in a slow rendering speed and excessive memory usage.

\section{Conclusion}
We proposed a scalable novel view synthesis pipeline that completes larger amounts of incomplete scene data with plausible information.
Our method builds upon a sparse grid-based feature representation that jointly encodes local geometry and texture information.
In contrast to NeRF or NSVF which requires test-time optimization, our pipeline directly encodes the few-shot input views in a feed-forward manner.
The scene embeddings with geometry and texture for the missing area are generated during completion in the 3D domain.
Compared with other approaches, our method effectively combines full 3D modeling which is crucial for spatial and temporal consistency with generative modeling.
\\

\boldparagraph{Acknowledgments.}
This work was supported by JSPS Postdoctoral Fellowships for Research in Japan (Strategic Program) and JSPS KAKENHI Grant Number JP20H04205. Z. Li was supported by the Swiss Data Science Center Fellowship program. Z. Cui was affiliated with the State Key Lab of CAD \& CG, Zhejiang University. M. R. Oswald was supported by a FIFA research grant.

\clearpage
%
%
\bibliographystyle{splncs04}
\bibliography{egbib}
\end{document}